\documentclass{article}
 \usepackage[dblblindworkshop, final]{tagds_2025_corrected}

\workshoptitle{Topology, Algebra, and Geometry in Data Science }

\usepackage[utf8]{inputenc} %
\usepackage[T1]{fontenc}    %

\usepackage{url}            %
\usepackage{booktabs}       %
\usepackage{amsfonts,amsmath,amssymb}       %
\usepackage{nicefrac}       %
\usepackage{microtype}      %
\usepackage{xcolor}         %
\usepackage{wrapfig}
\usepackage{multirow}
\usepackage{amsthm, wasysym}
\usepackage{svg}
\usepackage{booktabs, tabularx, caption}

\usepackage{color}
\usepackage{amscd, tikz-cd, mathrsfs, tikz}
\usepackage{todonotes}
\usepackage[font={it,small},labelfont={bf,it,small}]{caption}
\usepackage{subcaption}
\usepackage{float}
\usepackage{enumitem}
\setlist[itemize]{topsep=0pt,itemsep=0pt,partopsep=0pt}
\setlist[enumerate]{topsep=0pt,itemsep=0pt,partopsep=0pt}
\usepackage[pdftex,colorlinks=true, pdfstartview=FitV, linkcolor=magenta, citecolor=blue, urlcolor=blue]{hyperref}

\theoremstyle{plain}

\theoremstyle{definition}

   \newtheorem{question}{Question}

\usepackage[utf8]{inputenc}
\usepackage{booktabs,tabularx,caption,array}

\title{Topological Signatures of ReLU Neural Network Activation Patterns}

\author{
    Vicente~Bosca\thanks{These authors have equal contribution.} \\
    Applied Math and Computational Sciences\\
    University of Pennsylvania \\
    Philadelphia, PA 19104 \\
    \texttt{vicenteg@sas.upenn.edu} \\
    \And
    Tatum~Rask$^*$ \\
    Department of Mathematics\\
    Colorado State University \\
    Fort Collins, CO 80523 \\
    \texttt{tatum.rask@colostate.edu} \\
    \And
    Sunia~Tanweer$^*$ \\
    Departments of Mechanical Engineering \& \\
    Computational Mathematics, \\Sciences and Engineering (CMSE)\\ 
    Michigan State University \\
    East Lansing, MI 48824 \\
    \texttt{tanweer1@msu.edu} \\
    \And
    Andrew R.~Tawfeek$^*$ \\
    Department of Mathematics \\
    University of Washington\\
    Seattle, WA 98195 \\
     \texttt{atawfeek@uw.edu} \\
    \And
    Branden~Stone \\
    Georgia Tech Research Institute \\
    Atlanta, GA 30332 \\
    \texttt{branden.stone@gtri.gatech.edu} \\
}

\begin{document}

\maketitle

\begin{abstract}

This paper explores the topological signatures of ReLU neural network activation patterns. We consider feedforward neural networks with ReLU activation functions and analyze the polytope decomposition of the feature space induced by the network. Mainly, we investigate how the Fiedler partition of the dual graph and show that it appears to correlate with the decision boundary---in the case of binary classification. Additionally, we compute the homology of the cellular decomposition---in a regression task---to draw similar patterns in behavior between the training loss and polyhedral cell-count, as the model is trained.

\end{abstract}

\section{Introduction}

In recent years, neural networks have revolutionized the ability to learn from data, driving innovation across various fields \cite{lecun2015deep,li2021survey}. Despite significant study, much remains to be understood about how these models learn from data and how to effectively measure their robustness. In effort to understand how these models learn, we focus on applying topological methods to these networks motivated by the following broad question. 
\begin{question}
    Can we identify topological features that correlate with the network's performance during training?
\end{question}

\begin{figure}[!htbp]
  \centering
  \includegraphics[width=0.6\linewidth]{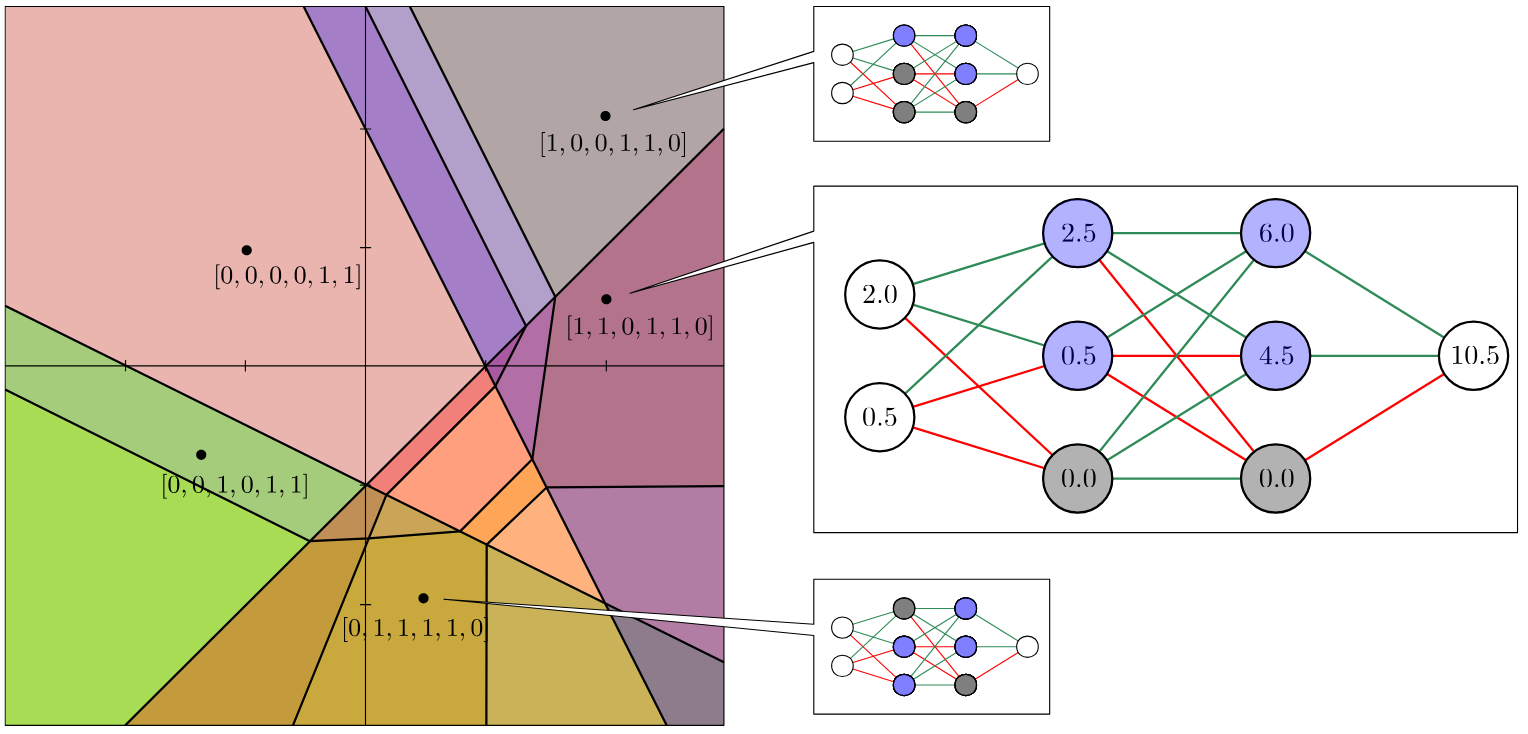}
  \caption{Example of polyhedral decomposition of the domain of a FFRNN. Grayed out activation function outputs correspond to ReLU value of zero at that neuron, indicating a 0 entry in the binary state vector.}
  \label{fig:polydecomp}
\end{figure}

This question has been investigated from various perspectives. For instance, \cite{naitzat2020topology} studies the shape of data as it passes through a ReLU network, and similarly \cite{Wheeler2021Landscapes} investigates this evolving topology of data to evaluate network performance. Others aim to use topological features to guide the model during training \cite{pmlr-v108-gabrielsson20a}. In this paper, we focus on the polyhedral decomposition of the input space $\mathbb R^n$ of the neural network, as defined in \cite{liu2023relu} (see Figure~\ref{fig:polydecomp}). We investigate topological signatures of performance in feedforward neural networks for classification and regression tasks. Our key findings, based on experiments with the input space $\mathbb R^2$, are:
\begin{itemize}
\item The \textit{weighted} Fiedler partition of the graph appears to correlate with the decision boundary for binary classification when the network exhibits grokking.
\item Moments of training instability are not merely numerical artifacts, but are associated with a deeper, transient reorganization of the network's internal topological representation.
\end{itemize}

The remainder of this paper is organized as follows. In Section~\ref{sec:dual_graph}, we consider classification tasks and examine the dual graph of the polyhedral decomposition. Here, the weighted Fiedler partition of the graph is defined by the eigenvector corresponding to the smallest non-zero eigenvalue. Two experiments demonstrate the correlation with the decision boundary for binary classification when the network exhibits grokking \cite{humayun2024deep,power2022grokking}.

In Section~\ref{sec:betti}, we analyze regression tasks and study the topological evolution of the cell complex by training a model, computing the corresponding polyhedral decomposition of the input space, and applying a random filtration to the cell complex. This allows us to study the evolution of the homology and cell count across various epochs. We find a correlation between the training loss of the model and the Betti numbers of the filtration, which similarly translates to a correlation between the $f$-vector (cell count) and loss. Specifically, we consider a physics-informed neural network modeling a Duffing oscillator (Appendix~\ref{secapp:physics}), a canonical nonlinear dynamical system.

\subsection{Preliminaries}\label{sec:background}

The main object of study in the work is the $(L+1)-$layer feedforward ReLU neural network (FFRNN) (See Appendix~\ref{secapp:ffrnn}):
\begin{multline}
    \label{eq:nn}
\mathbb{R}^m\xrightarrow[\text{ReLU}]{(W_1, b_1)}\mathbb{R}^{h_1}\xrightarrow[\text{ReLU}]{(W_2, b_2)}\mathbb{R}^{h_2}\rightarrow\ldots\rightarrow\mathbb{R}^{h_{L-1}}
\xrightarrow[\text{ReLU}]{(W_{L}, b_{L})}\mathbb{R}^{h_L}
\xrightarrow[]{(W_{{L+1}}, b_{{L+1}})}\mathbb{R}^{n}.
\end{multline}

In this model, $\mathbb{R}^m$ is the input space, $\mathbb{R}^n$ is the output space, and $h_i$ corresponds to the number of nodes in layer $i$. In other words, this network has architecture $(h_0=m, h_1,h_2,\dots,h_L,h_{L+1}=n)$.

As detailed in \cite{liu2023relu}, FFRNNs inherently define {\it binary state vectors} that create a polyhedral decomposition of the domain (Figure~\ref{fig:polydecomp}). More explicitly, consider the model~\eqref{eq:nn} above. Given an input data point $x\in\mathbb{R}^m$ in the input space, we have for each hidden layer $i$ (so $1\leq i\leq L)$ a binary (bit) indicating whether that neuron is ``on'' or ``off,'' which we may assemble into a vector \[s_i(x)=[s_{i,1}(x) \ \ldots\  s_{i,h_i}(x)]^{\top}\in\mathbb{R}^{h_i},\] where $s_{i, j}(x)$ (with $1\leq j\leq h_i$) is defined as follows:
\begin{equation}
\label{eq:bitvector}
        s_{i, j}(x)=  
    \begin{cases}
      1 & \text{if}\  w_{i,j}F_{i-1}(x)+b_{i,j}>0 \\
      0 & \text{if}\  w_{i,j}F_{i-1}(x)+b_{i,j}\leq 0.
    \end{cases} 
\end{equation}
Thus, for each point $x\in \mathbb R^m$ in the input space, we have a sequence of binary vectors $s_1(x), s_2(x), \dots, s_L(x)$.
We can stack the binary vectors associated to $x$  to make a long column vector 
\begin{equation}
\label{eq:bitvec}
    s(x)= [s_1^{\top}(x)\  \ldots \ s_{L}^{\top}(x)]^{\top}\in\mathbb{R}^{h},
\end{equation}
where $h = \sum_{i=1}^L h_i$ is the total number of nodes in the hidden layers. We call $s(x)$  the binary vector of $x$. These binary vectors create a polyhedral decomposition in the input space such that all regions associated with the same binary vector form one polytope (Figure~\ref{fig:polydecomp}). 

\section{Classification on the dual graph}\label{sec:dual_graph}
In Liu et al \cite{liu2023relu}, the authors utilize the \textit{dual graph} of the decomposition induced by the bit vectors to run persistent homology. Inspired by their work, we propose this dual graph as a useful tool for studying the topology of a neural network. Namely, we will use (a weighted version of) the \textit{graph Laplacian} \cite{KirchhoffUeberDA, Fiedler1973, Anderson01101985} to analyze the geometry of neural networks for classification problems.

We are not the first to use discrete Laplacians to analyze and study neural networks and their use in classification tasks. 
For example, \cite{Jamil_2023_CVPR} applies the Fiedler vector partition to ``representational dissimilarity matrices''. However, to our knowledge, we are the first to consider the graph Laplacian in this specific setting.

\subsection{Graph Laplacian}

Let $G = (E, V)$ be a graph with vertex set $V = \{v_j\}_j$ and edge set $E \subseteq V \times V$. First, construct the coboundary matrix $\partial: \mathbb{R}^{|V|} \to \mathbb{R}^{|E|}$ as follows: suppose edge $i$ is given by $e_i = (s,t)$. Then, $\partial(i, j) = [v_j: e_i]$, where 
\[
[v_j:e_i] = \begin{cases}
1 & \text{if } v_j = t,\\
-1 &\text{if } v_j = s,\\
0 & \text{otherwise.}
\end{cases}
\]
Then, the \textit{graph Laplacian} is defined as $L = \partial^T \partial$ \cite{Anderson01101985}.

There is vast literature on the graph Laplacian and its theory and uses in applications; however, for the purpose of this work, we will focus on certain spectral properties of the graph (see \cite{Cvetković_Rowlinson_Simić_2009} for a more complete discussion). For one, $\dim(\ker L)$ counts the number of connected components of $G$. In our case, $\dim(\ker L) = 1$. The smallest non-zero eigenvalue, call it $\lambda_1$, is often called the \textit{algebraic connectivity} of $G$ \cite{Fiedler1973}. The associated eigenvector, called the \textit{Fiedler vector}, has entries that are guaranteed to sum to 0. This makes the Fiedler vector easily amenable to binary partitions: nodes are classified based on whether they are assigned a positive or negative value. It is known that this partition approximates a minimum cut on a graph \cite[Equation 7.12]{Cvetković_Rowlinson_Simić_2009}. For shorthand, we will refer to this partition as the \textit{Fiedler partition}.

The definition of the graph Laplacian above implicitly assumes that our vertex and edge sets form orthonormal bases. By changing the inner product on our edge and vertex spaces, we inherently change the Laplacian. Specifically, we can \textit{weight} our edges and vertices to produce a new inner product. Denote by $W_V$ the diagonal matrix where $W_V(i,i)$ is given by the weight of the $i$th vertex. Construct $W_E$ similarly. Then, the weighted Laplacian is given by $L = W_V^{-1} \partial^T W_E \partial$.

\subsection{Dual graph for polyhedral decomposition}

Now, we will construct a graph $G = (V, E)$ from a FFRNN with architecture given in Equation~\ref{eq:nn}. Let $V = \{s(x)\}$ obtained from Equation~\ref{eq:bitvec}, and let $E$ be given by pairs of vertices (i.e. binary vectors) that differ in exactly one entry. That is, pairs of binary vectors that have \textit{Hamming distance} of 1 \cite{liu2023relu}. See Figure~\ref{fig:toy} for an illustration of the dual graph. We aim to answer the question: 

\begin{question}
    Does the Fiedler vector partition the dual graph to reflect the decision boundary of a binary classification problem? If not, can we adjust the weights so that it does?%
\end{question}

In summary, we found that the partition of the dual graph by the Fiedler vector from the unweighted Laplacian is not accurate. In the paragraphs to follow, we propose a choice of vertex weights so that the Fiedler partition from the \textit{weighted} Laplacian more accurately reflects class labels for binary classification tasks.%

Place the following weights on the nodes of the dual graph $G$: let $W_V(i,i)$ equal one plus the number of training data points contained in polytope $i$ (we use a function in the \texttt{GoL\_Toolbox} to obtain this count \cite{Liu_GoLPaper, CalgarGoL}). Let $W_E$ be the identity matrix. Experimental results on small examples (see Section~\ref{subsec:experiments}) demonstrate that this choice of weights will, indeed, reflect the decision boundary of a network trained on a binary classification problem \textit{that demonstrates grokking}. Grokking, also known as \textit{delayed generalization} \cite{power2022grokking}, geometrically manifests as an increase in the polytope regions around training points once the network reaches negligible training error. At the same time, there is a greater concentration of (smaller) polytopes around decision boundaries \cite{humayun2024deep}. These smaller polytopes concentrated near the boundary will, then, be assigned a smaller weight in the dual graph than those polytopes centered near training data. Thus, we expect that the weighted Fiedler vector will ``cut'' near the decision boundary. 

Although our small experiments show success, we have yet to prove that this weighted Laplacian does, indeed, approximate a (weighted) minimum cut. Future work includes proving such a statement directly or utilizing existing results for edge-weighted \cite{Fiedler1975} and vertex-weighted \cite{Xu_2020} Laplacians.

\subsubsection{Experiments}
\label{subsec:experiments}
We ran two small experiments, following the same basic steps, to test the weighted Fiedler partition.

\begin{enumerate}
\item Train a FFRNN with varying architectures and epoch counts to perform binary classification. 
\item Compute the polytope decomposition of the input space using \verb|GoL_Toolbox| \cite{CalgarGoL}.
\item Compute and visualize the dual graph of the polytope decomposition.
\item Calculate the Fiedler partitions for both the unweighted Laplacian and the training point-weighted Laplacian.
\item Evaluate the accuracy of the Fiedler partition by:
\begin{enumerate}
\item Computing the average class label within each polytope.
\item Assessing accuracy via (1) proportion of polytopes with incorrect class identification, and (2) $L_2$ error (defined as the $L_2$ norm of the vector of differences between average class labels and predicted values).
\end{enumerate}
\end{enumerate}

\begin{table}%
\small
\centering
\caption{Experimental results testing the weighted and unweighted Fiedler partitions. The signs of the entries of the Fiedler vector are used to predict class labels. Zero misclassification and $L_2$ error means that the Fiedler partition perfectly matches the class labels.  \smallskip}
\begin{tabular}{cccccccc}
    \toprule
     & & \multicolumn{2}{c}{Loss} & \multicolumn{2}{c}{Unweighted} & \multicolumn{2}{c}{Weighted} \\
    \cmidrule(lr){3-4}\cmidrule(lr){5-6}\cmidrule(lr){7-8}
    Dataset & Architecture & Train & Test  & Missclass. (\%) & $L_2$ Error & Missclass. (\%) & $L_2$ Error \\
    \midrule
    Circles & $(2,6,6,2)$ & 0.00002 & 0.00016 & 19.05\% & 2 & 0\% & 0 \\
    Moons & $(2,5,5,5,2)$ & 0.00002 & 0.00001 & 10\% & 1.20 & 0\% & 0.34 \\
    \bottomrule
\end{tabular}
\label{tab:graph_results}
\end{table}

\begin{figure}%
\centering
        \centering
        \includegraphics[width=0.21\textwidth]{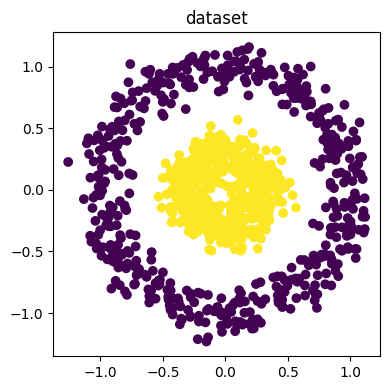}
        \includegraphics[width=0.22\textwidth]{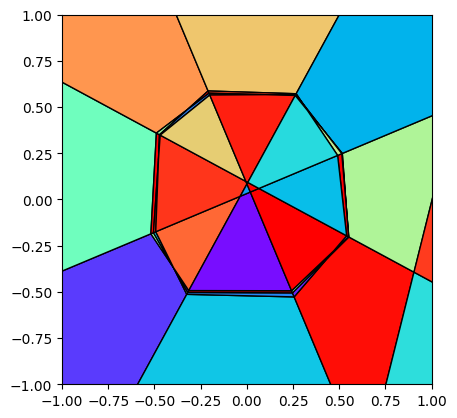}
        \includegraphics[width=0.275\textwidth]{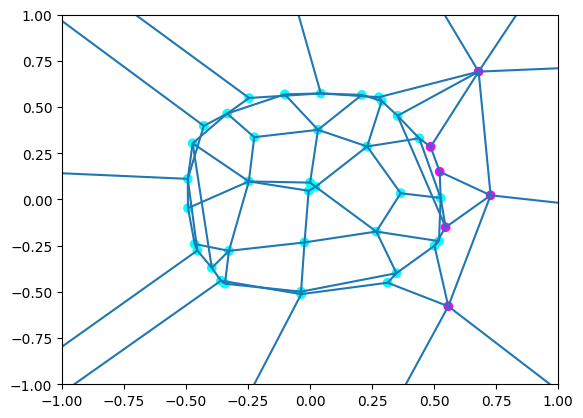}
        \includegraphics[width=0.275\textwidth]{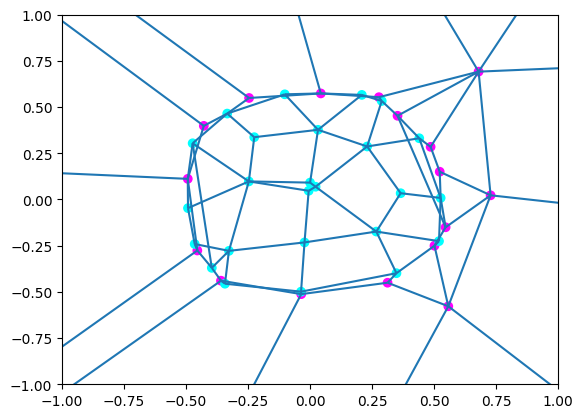} \\
        \includegraphics[width=0.21\textwidth]{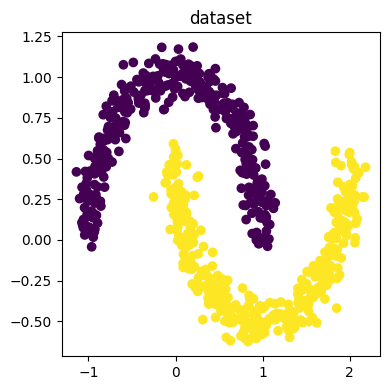}
        \includegraphics[width=0.27\textwidth]{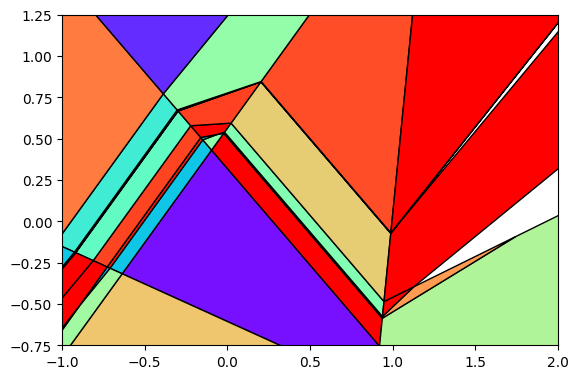}
        \includegraphics[width=0.245\textwidth]{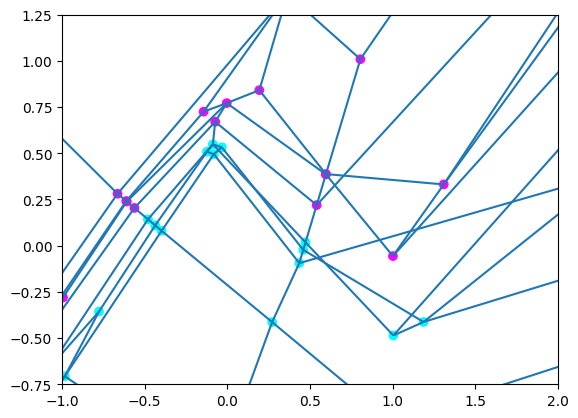}
        \includegraphics[width=0.245\textwidth]{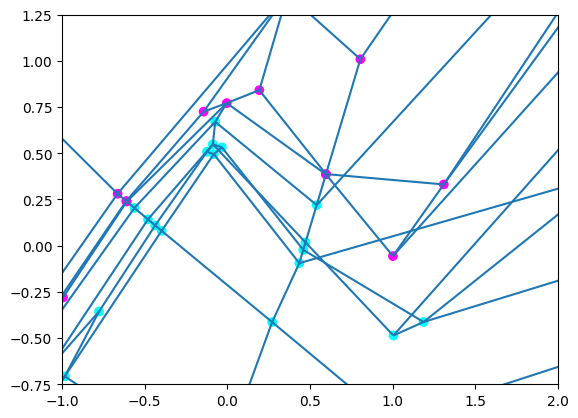}
    \caption{The two circles and two moons datasets (left column), their respective polytope decompositions (middle left), and the partitions of the dual graph by the unweighted (middle right) and weighted Fiedler vectors (right), indicated by the blue and purple points. Minimal loss and concentration along decision boundary demonstrates grokking behavior. }
    \label{fig:twocircles_experiment}
\end{figure}

\paragraph{Two circles experiment} As seen in Figure~\ref{fig:twocircles_experiment}, our Two Circles data contains points sampled along two concentric circle. We trained a FFRNN with 2 hidden layers (both of width 6) to learn the two circles binary classification task to 4000 epochs. As shown in Table~\ref{tab:graph_results}, model obtained 100\% accuracy on both the training and test datasets, with minimal loss on both training and testing data. The unweighted Fiedler vector missclassified 19.05\% of the polytopes with an $L_2$ error of 2. On the other hand, the weighted Fiedler vector missclassified 0\% of the polytopes with an $L_2$ error of 0. 

\paragraph{Two moons experiment} Figure~\ref{fig:twocircles_experiment} outlines the experiment with data sampled along two distinct arcs. Here, we trained a FFRNN with 3 hidden layers (all of width 5) to learn the two moons binary classification task to 2000 epochs. As expected, the model obtained 100\% accuracy on both the training and test datasets with minimal loss. The unweighted Fiedler vector missclassified 10\% of the polytopes with an $L_2$ error of 1.20. On the other hand, the weighted Fiedler vector missclassified 0\% of the polytopes with an $L_2$ error of 0.34. 

Because of the success of these small examples, we believe that the weighted Fiedler partition serves as a decent proxy for whether or not a network has achieved grokking.
That is, the dual graph and Fiedler partition offer an approach for analyzing decision boundaries and classification structure on well-trained networks; for complementary insights into the temporal evolution of network geometry, we next explore homological analysis of the full cell complex.

\section{Topological analysis via cell complex homology}\label{sec:betti}

Beyond the dual graph's focus on polytope adjacencies, we can alternatively analyze the full polyhedral decomposition as a cell complex (Appendix~\ref{secapp:cell}). This approach captures the hierarchical relationships between polytopes and their boundaries, enabling us to apply homological tools to track topological features that emerge and persist during training.

\subsection{Computing homology via random filtration}

To analyze the topological evolution of this cell complex during training, we compute the Betti numbers $\beta_i$ \cite{Hatcher} of the polyhedral decomposition at each training epoch. $\beta_0$ counts connected components, $\beta_1$ counts loops, $\beta_2$ counts voids, and higher-dimensional Betti numbers capture corresponding topological features. The examples presented here consider a bounded subset of $\mathbb{R}^2$ as our input space, so only $\beta_0$ and $\beta_1$ are non-trivial.

\begin{figure}%
    \centering
    \includegraphics[width=\textwidth]{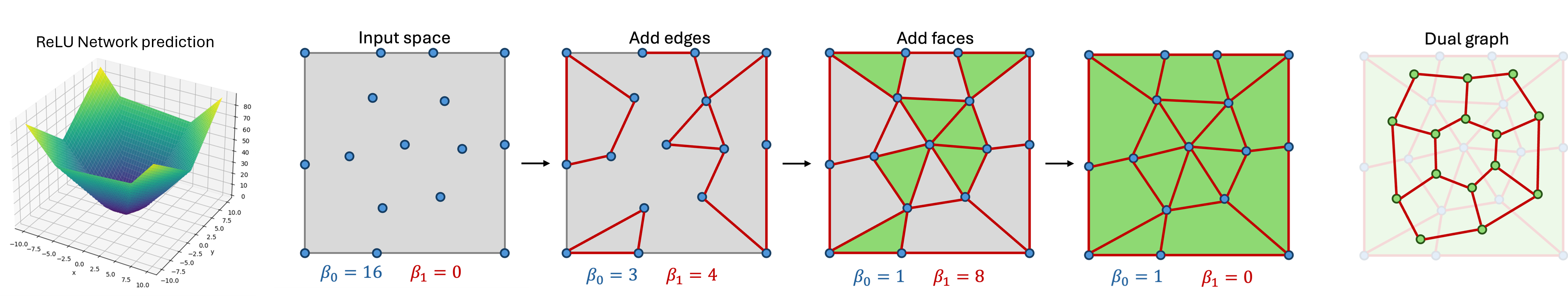}  %
    \caption{(Left) Prediction of a FFRNN trained to learn a paraboloid surface. (Middle) Snapshots of a random filtration on the primal complex of the polyhedral decomposition of such a network with their corresponding Betti numbers. First, randomly add all the vertices of the decomposition (blue), then add the edges (red, cycles are born), and finally add the faces (green, cycles are closed). (Right) Dual graph of the polyhedral decomposition: each polytope is now a vertex, and join two vertices if their corresponding polytopes share a face.}
    \label{fig:toy}  %
\end{figure}

If we were to compute the Betti 
numbers of the polyhedral decomposition of the input space directly, we would just recover the Betti numbers of the input space (for a 2D square $\beta_0=1$ and $\beta_1 = 0$). To reveal a deeper topological structure, we add cells progressively and compute the Betti numbers at each step. The way we add these cells is called a \textit{filtration}, a sequence of nested subcomplexes $K_0 \subseteq K_1 \subseteq \cdots \subseteq K_T$ with $K_0 = \emptyset$ and $K_T = K$. We then compute the \textit{Betti curves} as $\beta_i(t) = \dim H_i(K_t)$, for $i=0,1$ and $t \in [0,T]$, where $t$ denotes the stage of our filtration.  Specifically, we implement a \textit{random filtration} approach (see Figure~\ref{fig:toy} as an example):
\begin{enumerate}
    \item Begin adding the 0-cells (vertices) in the complex, selected uniformly at random
    \item At each timestep, randomly select and add one $1$-cell (edge) from the remaining edges
    \item Once all $1$-cells are added, proceed to randomly add $2$-cells (faces)
    \item Continue this process for higher-dimensional cells
\end{enumerate}
This filtration allows us to track how topological features emerge and persist as we reconstruct the complex. As edges are added, cycles begin to form (increasing $\beta_1$), and different connected components start to merge (decreasing $\beta_0$), which are subsequently filled when the corresponding faces are added (decreasing $\beta_1$). The Betti curves for this process are shown in Figure~\ref{fig:bettis_simple}.

To ensure robustness against the randomness in cell ordering, we perform multiple random trials and average the resulting Betti curves. The filtration parameter is expressed as the percentage of total cells added, enabling comparison across different epochs despite varying complex sizes.

Given the computational complexity of directly computing homology from boundary matrices (which can involve thousands of cells per dimension for medium-sized networks), we employ a multi-step computational pipeline. First, we use edge subdivision \cite{berzins2023polyhedralcomplexextractionrelu} to compute the sign vectors of the polyhedral decomposition. Then, we apply the perturbation method described in the same work to construct the full cell complex structure from these sign vectors. Finally, we use the \texttt{PHAT} package \cite{phat} to efficiently compute the homology with $\mathbb{Z}_2$ coefficients of the resulting complex under our chosen filtration.

\subsection{Visualizing topological evolution during training}

For the following results, we used a regression problem from a Physics-Informed Neural Network (PINN); the details can be found in Appendix~\ref{secapp:physics}. For each training epoch, we extract the polyhedral decomposition and compute averaged Betti curves, which we visualize in two complementary ways to reveal different aspects of the topological evolution.\footnote{These experiments were conducted on a system equipped with an Intel Xeon Silver 4214R CPU, running at a base clock speed of 2.40 GHz. The system has a total of 24 CPUs having 700 KB L1 cache, 12 MB L2 cache, and the system as a whole has 16.5 MB of shared L3 cache.}.

\begin{figure}%
    \centering
    \includegraphics[width=\textwidth]{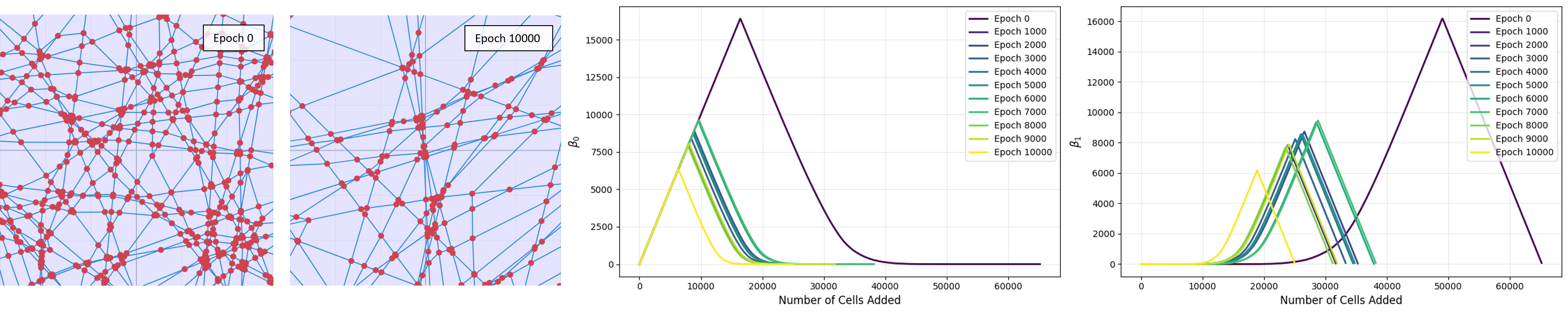}  %
    \caption{(Left) Snapshots  (zoomed in to the square $[0.3,0.3]^2$) of the polyhedral decomposition of a PINN at the beginning  (epoch 0) and end of training (epoch 10000). In red are the vertices (0-cells), in solid blue the edges ($1$-cells), and in a lighter blue the faces ($2$-cells). More cells are observed at the beginning of training. This number seems to decrease as we train, with the emergence of some structure. (Right)  Corresponding average Betti curves, $\beta_0$ and $\beta_1$, of each stage of training for a random filtration of the resulting complex.} %
    \label{fig:bettis_simple}  %
\end{figure}

\paragraph{Betti curves}
Figure~\ref{fig:bettis_simple} shows the average Betti curves themselves, with the $x$-axis representing the number of cells added (across all dimensions) and the $y$-axis showing the Betti number value. For $\beta_0$ in 2D input spaces, the curve begins at the origin and increases linearly (with slope $1$) as vertices are added, reaching its peak when all vertices are included. As edges are subsequently added, $\beta_0$ initially decreases approximately linearly since each edge typically merges two connected components, although this decrease slows as the complex becomes more connected, ultimately converging to $1$ (a single connected component).

The behavior of $\beta_1$ follows a complementary pattern. Starting at zero, it remains flat until sufficient edges create the first loops, then increases, eventually growing linearly as most new edges create additional cycles. The peak occurs when all edges are added, with the peak value equaling the number of polytopes that remain to be filled. As $2$-cells (faces) are added in 2D spaces, $\beta_1$ decreases linearly (with slope $-1$) until reaching zero, when all faces have filled the loops.

The peaks of our Betti curves directly reveal the \textit{f-vector} of the polyhedral decomposition, defined as $f(K) = (f_0(K), f_1(K), \ldots, f_{\dim K}(K))$ where $f_i(K)$ denotes the number of $i$-cells in $K$. Specifically, $\max(\beta_0) = f_0$ (the number of vertices) and $\max(\beta_1) = f_2$ (the number of $2$-cells) when all edges but no faces have been added. This provides a computationally efficient alternative to full homology computation: while computing Betti numbers becomes intractable for networks with thousands of cells per dimension, the $f$-vector can be extracted directly from the cell complex construction. Note that the $f$-vector components are constrained by the Euler characteristic of the domain space. For our 2D square, $\chi = f_0 - f_1 + f_2 = 1$, which together with the fact that $f_i\geq 0$ by definition implies that $f_0\sim f_2\sim\frac{1}{2}f_1$, which is experimentally verified in Figure~\ref{fig:fvector}.

\begin{figure}%
    \centering
    \includegraphics[width=0.82\textwidth]{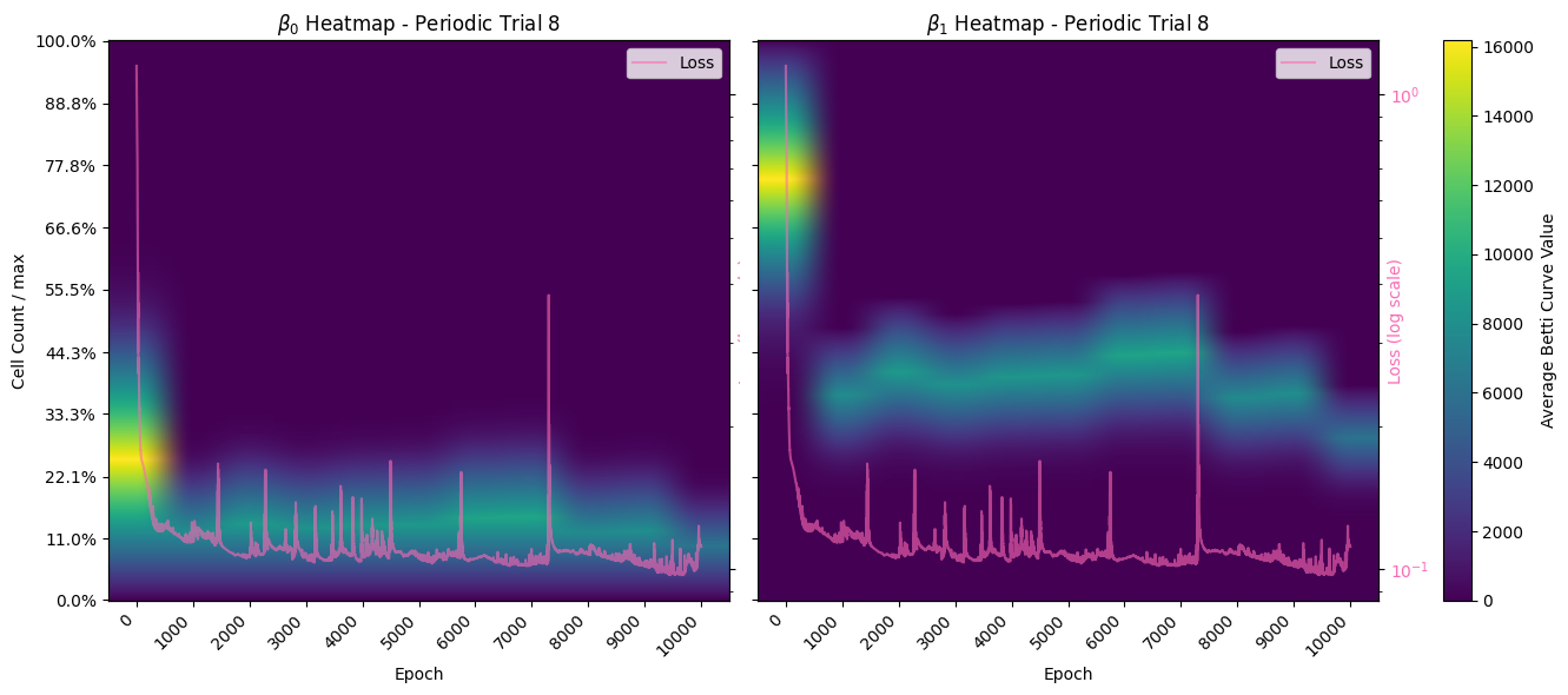}  %
    \caption{Heat maps of the Betti curves corresponding to a trained PINN, with the training loss overlaid. (Left) For $\beta_0$ we observe smoother changes. (Right) For $\beta_1$, the changes are sharper and correlate with substantial changes in the training loss function.}
    \label{fig:heat}  %
\end{figure}

\paragraph{Heat maps}
The second visualization (Figure~\ref{fig:heat}) normalizes and compresses this topological information into heat maps, easing comparison across training epochs. In these heat maps, the $x$-axis represents training epochs, the $y$-axis shows the number of cells added divided by the maximum number of cells across all epochs (usually achieved at the early stages of training). The color encodes the average Betti curve value at that filtration percentage. Each vertical slice of the heat map essentially represents a compressed version of the full Betti curve from that epoch, with color encoding what would be the $y$-axis value in the Betti curve plot. We then stack them next to each other to visualize how the values change relative to the maximum number of cells for all epochs.

\subsection{Geometric Reorganization and Correlation between topological complexity and loss}

\begin{figure}%
    \centering
    \includegraphics[width=\textwidth]{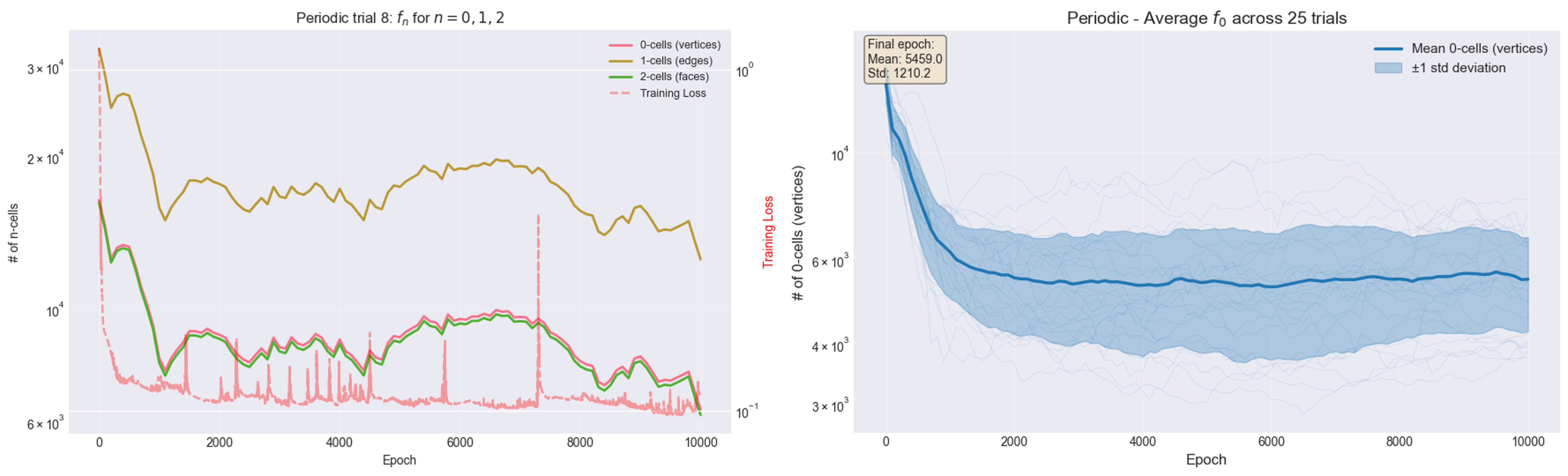}  %
    \caption{(Left) The three components of the $f$-vector across training epochs for a trained PINN with the training loss overlaid. Spikes in the training loss lead to changes in those quantities. (Right) The average $f_0$ (\# of $0$-cells) for the 25 trials of training a neural network using the periodic dynamical system. Overall, the number of cells decreases along training.}
    \label{fig:fvector}  %
\end{figure}

During the training process, while a general trend of decreasing loss was observed, characteristic ``spiking'' behavior can be noted in the training loss curve, see Figure~\ref{fig:fvector} (left). This phenomenon, where the loss exhibits sudden, transient increases before returning to a downward trend, is not uncommon in neural network training, and is known to be connected to complexities in the loss landscape, training dynamics and the edge-of-stability phenomenon~\cite{spiking_loss}.

A key finding from our analysis of the training dynamics relates the topological properties of the polyhedral decomposition with the loss. We computed Betti vectors for the decomposition at each training epoch. The filtration parameter for this topological analysis was defined as the percentage of randomly added complexes, providing a measure of the network's connectivity and structure. A notable observation was the correlation between the training loss and the filtration value at which the maximum Betti number $\beta_{\text{max}}$, was achieved. Specifically, we found that sudden ``spikes'' in the training loss corresponded to an increase in this critical filtration value (Figure~\ref{fig:heat}). This suggests that during epochs where the optimizer temporarily struggles to minimize the loss, the internal topological structure of the network becomes more complex, requiring a higher filtration threshold to reveal its most significant persistent homology features. The observed relationship indicates that moments of training instability are not merely numerical artifacts, but are associated with a deeper, transient reorganization of the network's internal representation. This result suggests that the loss landscape's complexity is directly reflected in the evolving homology of the network's decomposition, potentially allowing us to use the homology as a proxy for the loss without requiring access to training data---an endeavor left for future work.

A deeper examination of the heat maps in Figure~\ref{fig:heat} reveals a fundamental geometric reorganization throughout the training process. The downward shift of peak Betti values in early epochs---from approximately 25\% to 11\% for $\beta_0$ and from 75\% to 35\% for $\beta_1$---indicates substantial evolution in the number of cells across different dimensions. Considering only the max of $\beta_0$ and $\beta_1$ (brightest region in the plot), we see an initial decrease with respect to the early stages of training---reflecting a reduction in the total number of cells. Subsequent fluctuations align closely with abrupt variations in the training loss. This phenomenon is especially evident in Figure~\ref{fig:bettis_simple}. Immediately prior to the final big spike in loss, the number of cells tend to increase, followed by a decrease again. This is better captured in the left of Figure~\ref{fig:fvector}, where we plot the number of cells per epoch. The ``reorganizing'' behavior after a big training loss spike was observed consistently across trials. Nevertheless, the overarching trend for all trials (shown as average in right panel of Figure~\ref{fig:fvector}) is reduction in the total number of cells.

\textbf{Remark.}
The heat maps in Figure~\ref{fig:heat} normalize Betti curves by the maximum number of cells across all training epochs rather than within each epoch. For 2D bounded input spaces, the Euler characteristic constraint ($\chi = 1$) creates dependencies: approximately equal number of 0-cells and 2-cells, with roughly twice as many 1-cells. Under our random filtration, $\beta_0$ peaks at 25\% (all 0-cells added) and $\beta_1$ at 75\% (all 1-cells added), visible as horizontal bands in early epochs where the maximum cell count occurs. While this normalization effectively tracks the evolution of cell counts across dimensions due to these constraints, the heat maps capture more than just the $f$-vector evolution. For example, in the middle and left figures in Figure \ref{fig:bettis_simple} before 20,000 cells have been added, we observe non-trivial topological behavior where adding 1-cells does not immediately reduce the number of connected components, indicating complex geometric behavior beyond simple cell enumeration. A within-epoch normalization would yield uninformative horizontal lines at fixed percentages due to the 2D constraints, missing these subtle topological transitions.

To quantify the relationship between topology and training dynamics, we compute different statistical correlations, reported in Tables \ref{tab:main-correlations} and \ref{tab:appendix-correlations}. The analysis confirms a moderate-to-strong positive correlation between topological complexity and training loss. This correlation is robust to detrending, indicating that the relationship is not merely an artifact of shared temporal trends. The correlation of first differences further confirms that changes in topology coincide with changes in loss, strengthening the dynamic relationship interpretation. Cross-correlation analysis reveals the strongest relationship at a lag of +25 and +100 epochs for the chaotic and periodic dynamics systems, respectively, suggesting that geometric reorganization of the network's polyhedral decomposition \textit{precedes} shifts in the loss landscape. This temporal ordering not only supports the hypothesis that topological complexity serves as a leading indicator of optimization dynamics, but also highlights its potential as a predictive diagnostic for abrupt changes in training dynamics. This can prove to be very significant for understanding, monitoring and potentially controlling the learning process of neural networks.

\begin{table}%
\centering
\scriptsize
\caption{Main correlation results (means $\pm$ std) across $n=25$ trials. For each method we report mean correlation ($\bar r \pm$ std), mean $p$-value ($\bar p \pm$ std) and number of trials with $p<0.05$. ``Max cross-corr'' shows the lag (in epochs) at which the mean cross-correlation is maximal and the corresponding statistics.}
\label{tab:main-correlations}
\begin{tabularx}{\textwidth}{@{} l *{6}{>{\centering\arraybackslash}X} @{}}
\toprule
 & \multicolumn{3}{c}{\textbf{Chaotic} (n=25)} & \multicolumn{3}{c}{\textbf{Periodic} (n=25)} \\
\cmidrule(lr){2-4}\cmidrule(lr){5-7}
Method & $\bar r \pm \sigma$ & $\bar p \pm \sigma$ & Sig. & $\bar r \pm \sigma$ & $\bar p \pm \sigma$ & Sig. \\
\midrule

Raw Pearson (Loss vs 0-cells)
  & $0.69 \pm 0.13$ & $<10^{-6}$ & 25/25
  & $0.61 \pm 0.09$ & $<10^{-6}$ & 25/25 \\

Detrended (linear)
  & $0.82 \pm 0.11$ & $<10^{-6}$ & 25/25
  & $0.61 \pm 0.10$ & $<10^{-6}$ & 25/25 \\

Max cross-corr (lag)
  & \shortstack{lag = 25 \\ $0.70 \pm 0.13$}
    & \shortstack{$<10^{-6}$} & 25/25
  & \shortstack{lag = 100 \\ $0.74 \pm 0.14$}
    & \shortstack{$0.004 \pm 0.011$} & 25/25 \\
\bottomrule
\end{tabularx}

\end{table}

\section{Limitations and future directions}

\label{sec:limitations}

The major limitation of these methods is the computational complexity of the algorithms being employed. Scaling these tools to more complex neural architectures presents several challenges since the number of polytopes grows polynomially with network width and exponentially with depth \cite{Raghu2017, Hanin2019b}---while also scaling exponentially with input dimension. This exponential growth makes direct application of these methods computationally prohibitive for deep networks or high-dimensional inputs---with the current algorithms. 
Likewise, computing eigenvalues and eigenvectors of the graph Laplacian is also computational expensive. 

For future work, potential mitigation strategies could involve projecting the activation patterns onto principal components, thus reducing the effective dimensionality while preserving essential topological characteristics. Additionally, approximation algorithms for homology computation \cite{Otter2017} or sampling-based approaches could enable analysis of larger networks while sacrificing some precision in the topological measurements. Implementing algorithms and methods for quicker computation is also a worthwhile endeavour---for example, the TRACEMIN-Fiedler algorithm \cite{TRACEMIN-Fiedler} for Fiedler-vector computation. Further future work involves establishing theoretical justification of these empirical conclusions, exploring similar methods for multiclass classification problems, studying the topological invariance of the Euler characteristic, and investigating how the polyhedral decomposition evolves during training near decision boundaries using geometrically-informed filtrations, rather than the random filtrations employed in this work. In addition, as a reviewer noted, the relationship between grokking and larger polytopes suggests a potential link to circuit formation \cite{nanda2023progress}, wherein larger polytopes correspond to ``cleaner'', less complex circuits.

Another promising direction for future endevers involves extending our topological analysis to classification tasks. Recent work, \cite{masden2025algorithmic} demonstrates that the dual complex of ReLU networks forms a cubical complex, enabling the use of GUDHI. Building on this framework, we can investigate how the polyhedral decomposition evolves during training near decision boundaries using geometrically-informed filtrations. By tracking polytope birth and death through ``boundary-aware'' persistent homology across training epochs, we may characterize topological signatures of learning phenomena such as grokking, where networks transition from memorization to generalization. This temporal topological analysis could inform training algorithms that explicitly consider geometric complexity during optimization. More about the scalability issues can be found in Appendix \ref{app:scalability}.

 \section{Summary}

In this work, we explored topological methods to analyze FFRNN behavior by examining the polyhedral decomposition of the input space $\mathbb R^2$. We demonstrated that the weighted Fiedler partition correlates with the decision boundary during grokking and that Betti numbers of the filtration correlate with training loss. These findings emphasize the importance of geometric and topological structure over purely algebraic properties, potentially informing more effective neural network architectures and training methods.

\ack{This work was supported in part by the Georgia Tech Research Institute (GTRI) through their summer research internship program (GRIP). The authors thank GTRI for its support.}

\bibliographystyle{plain}
\bibliography{refs}

\appendix

\section{Feed Forward Networks}\label{secapp:ffrnn}

Consider an $(L+1)-$layer feedforward ReLU neural network (FFRNN): 
\begin{multline}
\mathbb{R}^m\xrightarrow[\text{ReLU}]{(W_1, b_1)}\mathbb{R}^{h_1}\xrightarrow[\text{ReLU}]{(W_2, b_2)}\mathbb{R}^{h_2}\rightarrow\ldots\rightarrow\mathbb{R}^{h_{L-1}}
\xrightarrow[\text{ReLU}]{(W_{L}, b_{L})}\mathbb{R}^{h_L}
\xrightarrow[]{(W_{{L+1}}, b_{{L+1}})}\mathbb{R}^{n}.
\end{multline}

In this model, $\mathbb{R}^m$ is the input space, $\mathbb{R}^n$ is the output space, and $h_i$ corresponds to the number of nodes at layer $i$. In other words, this network has architecture $(h_0=m, h_1,h_2,\dots,h_L,h_{L+1}=n)$.

We let $W_i\in \text{Mat}(h_i \times h_{i-1}, \  \mathbb{R})$ and  $b_i\in\mathbb{R}^{h_i}$ denote the weight matrix and bias vector of layer $i$, {respectively}. The activation function{s} for the hidden layers (layers $1, \dots, L$)  are assumed to be ReLU functions (applied coordinate-wise) while the map to the last layer (the output layer) is assumed to be affine linear (without a ReLU function being applied to the image). Recall that the ReLU function is the piecewise linear and continuous map $\text{ReLU}: \mathbb R \rightarrow \mathbb R$ given by 
\[
    \text{ReLU}(a)= \max\{0,a\}.
\]
It can be naturally extended to a piecewise linear continuous map on vector spaces (which we also denote as ReLU). More precisely, we define $\text{ReLU}: \mathbb R^{h_i} \rightarrow \mathbb R^{h_i}$, by applying the ReLU function to each coordinate of $x\in \mathbb R^{h_i}$. Given an input data point $x\in\mathbb{R}^m$, we denote the output of $x$ in layer $i$ as $F_i(x)$. Thus we have that $F_0(x) = x$ and 
\begin{equation}
    \begin{aligned}
\label{eq:outputderi}
  F_i(x)&=\text{ReLU}(W_iF_{i-1}(x)+b_{i})
    = \begin{bmatrix} \max\{0, w_{i,1}F_{i-1}(x)+b_{i,1}\}\\
    \vdots\\
    \max\{0, w_{i, h_i}F_{i-1}(x)+b_{i, h_i}\}
    \end{bmatrix}\in\mathbb{R}^{h_i},
\end{aligned}
\end{equation}
which expresses the value of each neuron on the $i$th layer of the network.

\section{Polyhedral Decomposition of Neural Networks}\label{secapp:nn_figure}

\begin{figure}[!htbp] 
    \centering
    \includesvg[width=\linewidth]{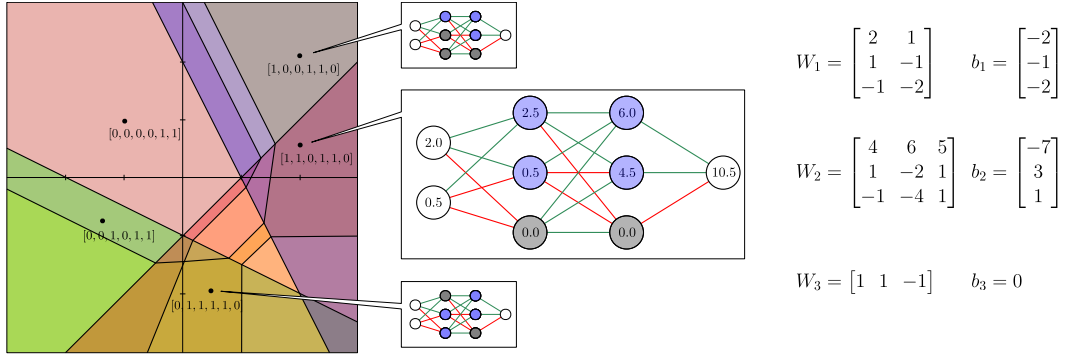} 
    \caption{The polyhedral decomposition of the input space $\mathbb{R}^2$ of a small FFNN having architecture $(2,3,3,1)$. Some of the binary vectors for various points in $\mathbb{R}^2$ are highlighted and their neural activation pattern within the network is illustrated.}
    \label{fig:ffnn}  
\end{figure}

\section{Cell Complex Structure}\label{secapp:cell}

The polyhedral decomposition of a FFRNN naturally induces a cell complex structure $K$ where:
\begin{itemize}
    \item $d$-dimensional cells are the polytopes themselves (for input dimension $d$),
    \item $(d-1)$-dimensional cells are the facets (boundaries between adjacent polytopes),
    \item lower-dimensional cells correspond to intersections of hyperplanes, and
    \item $0$-dimensional cells are the vertices where multiple hyperplanes meet.
\end{itemize}

This hierarchy can be encoded using \textit{sign vectors}, a generalization of the binary vectors introduced in Section~\ref{sec:background}. For a polytope $\mathcal{P}$, its sign vector is obtained by replacing zeros with $-1$ in the binary vector. Thus, a $-1$ in the sign vector means we are in the inactive part of the ReLU nonlinearity, and a $+1$ means we are in the active part. Apart from $-1$ and $+1$, the sign vector also has $0$s for those points in the input space that lie on the boundary of different regions. That is where the input of the ReLU maps is exactly zero, and corresponds to the lower-dimensional cells in the cell-complex. For full-dimensional polytopes, the sign vector only contains $-1$ or $+1$. For lower-dimensional cells, we introduce zeros at positions corresponding to neurons whose decision boundaries contain the cell. Specifically, a $(d-k)$-dimensional cell lying in the intersection of $k$ hyperplanes will have exactly $k$ zeros in its sign vector, with the remaining entries indicating which side of each hyperplane the cell lies on.

\subsection{Additional correlation measurements}

\begin{figure}[!htbp]  %
    \centering
    \includegraphics[width=\textwidth]{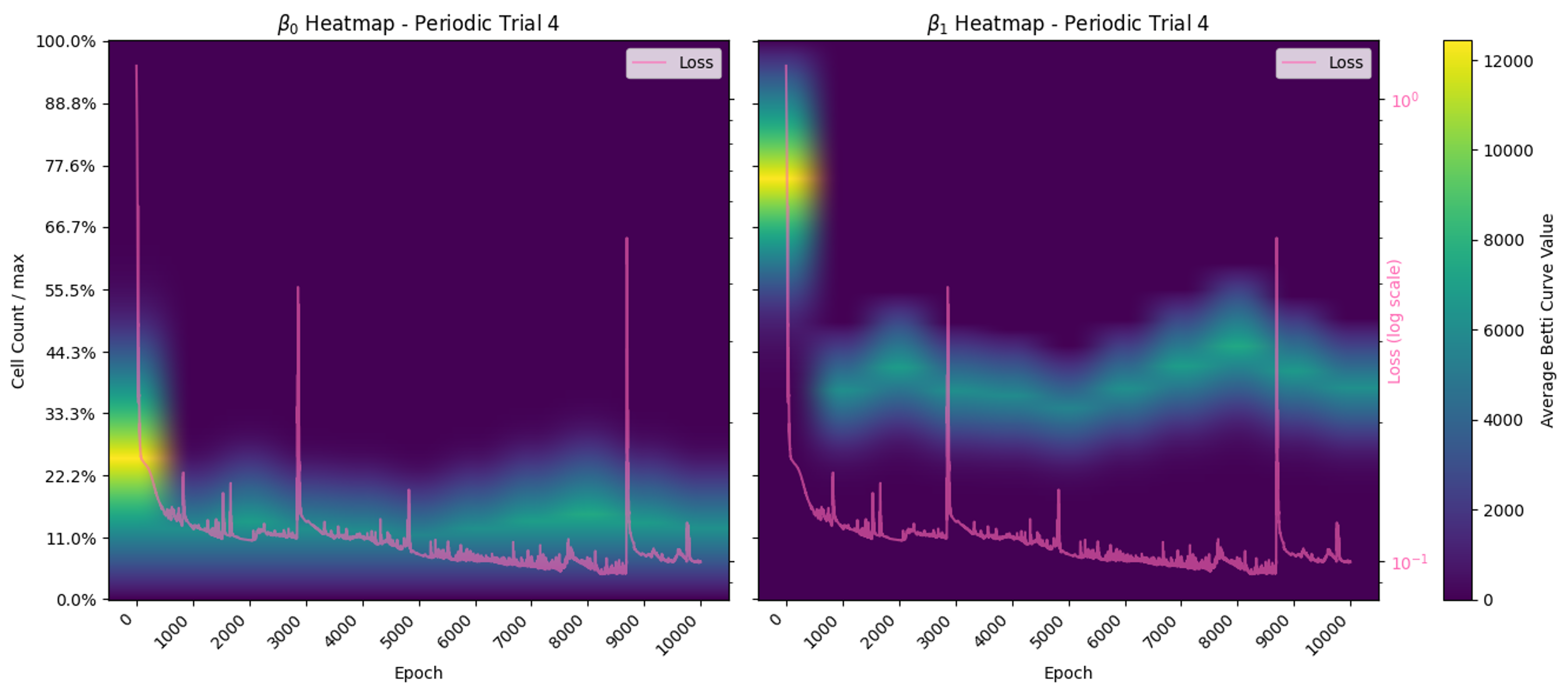}  %
    \caption{Heat maps of the betti curves for Trial 4 corresponding to a trained PINN, with the training loss overlaid. (Left) For $\beta_0$ we observe smoother changes. (Right) For $\beta_1$, the changes are sharper and correlate with substantial changes in the training loss function.}
\end{figure}

\begin{figure}[!htbp]  %
    \centering
    \includegraphics[width=\textwidth]{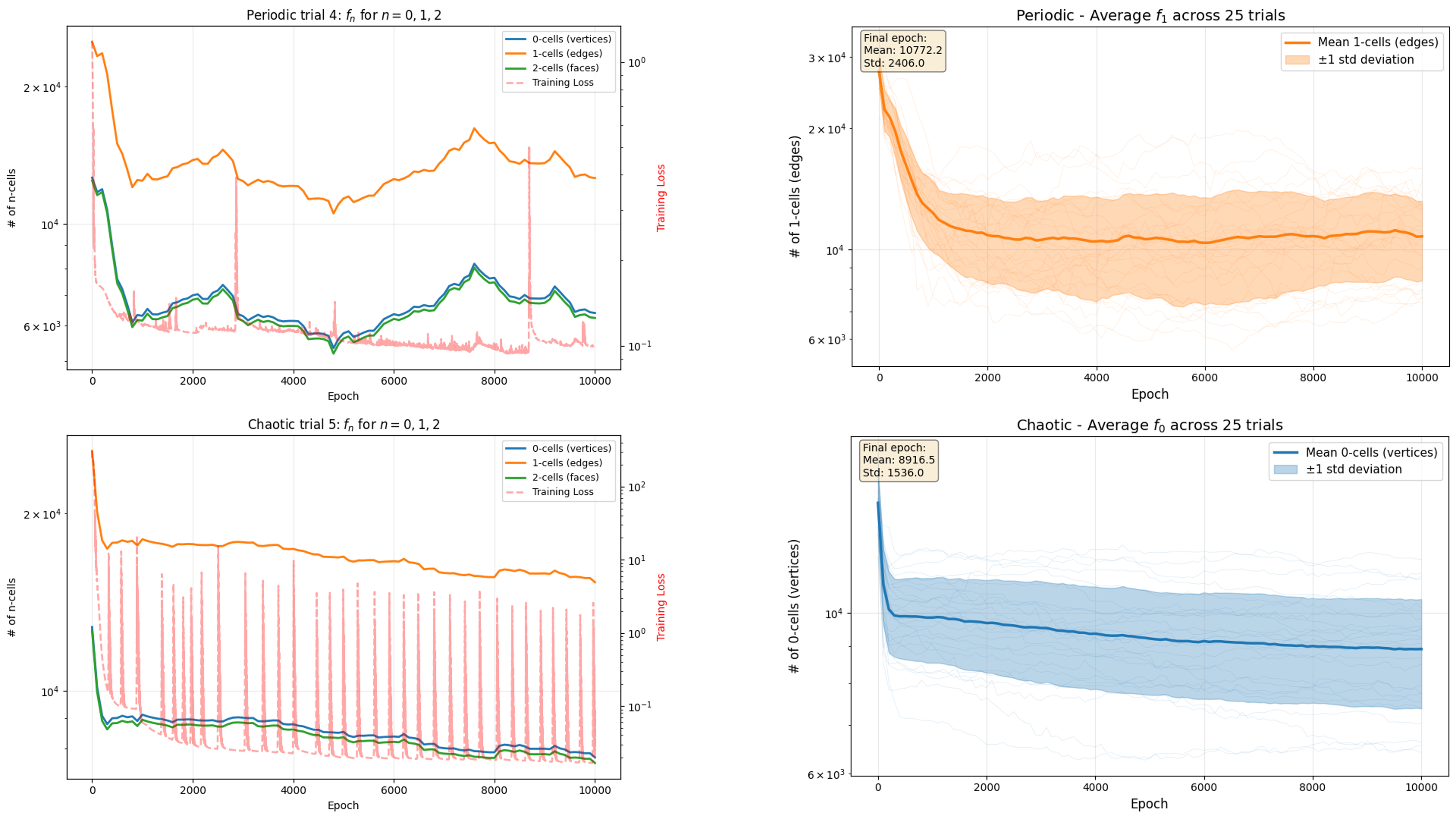}  %
    \caption{(Top left) The $f$-vector of another trial of the periodic dynamical system. (Top right) The averaged $f_1$ (instead of $f_0$) however these values are related through the Euler Characteristic. (Bottom left) is the $f$-vector of a chaotic dynamical system.  (Bottom right) The averaged $f_0$ for the 25 trials of the chaotic system.}
\end{figure}

\begin{table}[!htbp]
\label{table:app}
\centering
\scriptsize
\caption{Correlation robustness checks (First differences and Spearman rank). Values are means $\pm$ std across $n=25$ trials; ``Sig.'' counts trials with $p<0.05$.}
\label{tab:appendix-correlations}
\begin{tabularx}{\textwidth}{@{} l *{6}{>{\centering\arraybackslash}X} @{}}
\toprule
 & \multicolumn{3}{c}{\textbf{Chaotic} (n=25)} & \multicolumn{3}{c}{\textbf{Periodic} (n=25)} \\
\cmidrule(lr){2-4}\cmidrule(lr){5-7}
Method & $\bar r \pm \sigma$ & $\bar p \pm \sigma$ & Sig. & $\bar r \pm \sigma$ & $\bar p \pm \sigma$ & Sig. \\
\midrule

First differences
  & $0.88 \pm 0.14$ & $0.000 \pm 0.000$ & 25/25
  & $0.58 \pm 0.31$ & $0.038 \pm 0.128$ & 23/25 \\

Spearman rank
  & $0.29 \pm 0.20$ & $0.120 \pm 0.251$ & 20/25
  & $0.13 \pm 0.22$ & $0.205 \pm 0.290$ & 15/25 \\
\bottomrule
\end{tabularx}

\end{table}

\section{Physics-Informed Neural Network}
\label{secapp:physics}

We designed a Physics-Informed Neural Network (PINN)~\cite{Farea2024, PINNs} to learn the dynamics of the Duffing oscillator by predicting the system's displacement at the next time step, $\hat{x}(t+\Delta t)$, given the current time $t$ and current displacement $x(t)$. The network's architecture is a feedforward neural network such that: The input layer accepts a concatenated vector of $[t, x(t)]$. This input is then processed through a fully connected layer with ReLU activation, transforming the input into a higher-dimensional representation with $50$ neurons. 
Following this, the model incorporates three consecutive linear transformations followed by a ReLU activation. Finally, a linear output layer maps the output of the last hidden layer to a single scalar, representing the predicted displacement at the next time step, $\hat{x}(t+\Delta t)$. 

\subsection{Duffing Oscillator}

The Duffing oscillator is a canonical nonlinear dynamical system described by the second-order ordinary differential equation:

$$\frac{d^2x}{dt^2} + \delta \frac{dx}{dt} + \alpha x + \beta x^3 = \gamma \cos(\omega t)$$

\begin{figure}[!htbp]
    \centering
    \includegraphics[width=0.7\textwidth]{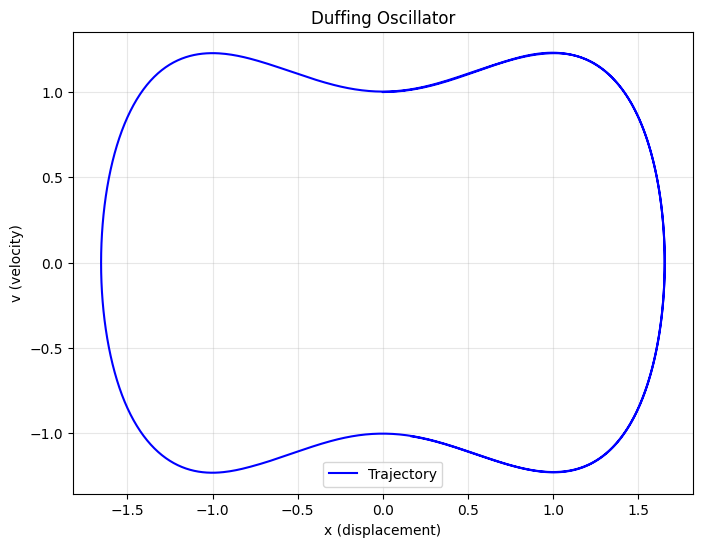}
    \caption{Phase space of a periodic duffing oscillator.}
    \label{fig:duffing}
\end{figure}

For our experiments, the parameters were set to $\delta = 0.0$, $\alpha = -1.0$, $\beta = 1.0$, $\gamma = 0.0$, and $\omega = 1.2$ to have a periodic regime, see Fig.~\ref{fig:duffing}. To generate training data, this equation was transformed into a system of first-order ODEs:
$$\frac{dx}{dt} = v \quad\quad \frac{dv}{dt} = - \alpha x - \beta x 3$$
This system was numerically solved with initial conditions $x(0) = 0$ and $v(0) = 1$ over a time interval of $t \in [0, 20]$. The resulting time series data for $x(t)$ served as the ground truth for training the PINN. %

\subsection{Physics-Informed Loss Function}

The training of the model relies on a composite loss function that combines both data-driven and physics-informed components. The total loss, $L_{total}$, is defined as the sum of the Mean Squared Error (MSE) of the data predictions ($L_{data}$) and the MSE of the residual of the governing differential equation ($L_{physics}$):
$$L_{total} = L_{data} + L_{physics}$$
The {data loss} measures the discrepancy between the network's predicted next state $\hat{x}(t+\Delta t)$ and the true next state $x(t+\Delta t)$ from the generated dataset:
$$L_{data} = \frac{1}{N} \sum_{i=1}^{N} (\hat{x}_i(t+\Delta t) - x_i(t+\Delta t))^2$$
where $N$ is the number of data points.
The {physics loss} enforces that the network's predictions follow the physical law of the oscillator's differential equation. This is achieved by computing the residual of the ODE using the network's outputs and their automatically differentiated derivatives. Let $\hat{x}_{pred}$ denote the network's prediction of $x(t+\Delta t)$. We obtain the approximate first derivative, $\hat{v}(t)$, using a finite difference:

$$\hat{v}(t) \approx \frac{\hat{x}_{pred} - x(t)}{\Delta t}$$
where $\Delta t$ is the time step between consecutive data points. The derivative of $\hat{x}_{pred}$ with respect to the input time $t$, denoted as $\frac{d\hat{x}_{pred}}{dt}$, is computed via automatic differentiation. The residual $R$ is then formulated by substituting these terms into the original Duffing equation. For our simplified Duffing equation, the residual is:
$$R = \frac{d\hat{x}_{pred}}{dt} - \left( -\delta \left(\frac{\hat{x}_{pred} - x(t)}{\Delta t}\right) - \alpha \hat{x}_{pred} - \beta (\hat{x}_{pred})^3\right)$$
The physics loss is the mean squared error of this residual:
$$L_{physics} = \frac{1}{N} \sum_{i=1}^{N} (R_i)^2$$
This combined loss function enables the PINN to learn both from the provided data and from the fundamental physical laws governing the Duffing oscillator.

\section{Scalability and Approximation Frameworks}
\label{app:scalability}

As discussed in Section \ref{sec:limitations}, the exact computation of the polyhedral decomposition for deep ReLU networks faces combinatorial explosion. The number of linear regions scales exponentially with the depth of the network and polynomially with the width \cite{Raghu2017}. Crucially, the complexity also scales exponentially with the dimension of the input space $m$ \cite{Raghu2017}, rendering exact enumeration intractable for high-dimensional data. To extend the topological analysis presented in this work to larger architectures, we propose shifting from exact computation to statistical approximation. Here we detail three specific methodologies to address the computational bottleneck.

\subsection{Sampling-Based Approximation}
The primary bottleneck in our current pipeline is the exhaustive traversal of the input space to identify every distinct activation region. We propose a \textit{Monte Carlo approach} where we sample $N$ points from the input domain and compute their binary codes to identify a subset of the active polytopes. We can then construct an approximate dual graph by connecting observed codes with a Hamming distance of 1. This reduction in graph size, scaling with the number of samples rather than the network's theoretical capacity, allows us to move beyond simple cell counting. By performing topological data analysis on this sampled subgraph, we can explore richer topological features and alternative filtrations that would be computationally prohibitive on the full exact complex.

\subsection{Dimensionality Reduction}
Given that the number of regions grows exponentially with the input dimension $m$ \cite{Raghu2017}, reducing the ambient space is a necessary step for scalability. A way to do this could be via \textit{PCA Projection}. Since real-world data often resides on lower-dimensional manifolds, we can project the input space onto the top $k$ principal components of the data (where $k \ll m$). The polyhedral decomposition is then computed on this $k$-dimensional affine subspace, avoiding the combinatorial explosion associated with the full ambient dimension.  More generally, we could resort to \textit{random slicing}. One can analyze the restriction of the neural network function to random 2D planes or 1D lines. Theoretical results in \cite{Hanin2019b} demonstrate that the number of regions along a 1D line provides a strictly bounded proxy for the global complexity. Extending this to 2D slices allows for the computation of Betti curves that reflect local topological complexity.

\subsection{Algebraic and Combinatorial Invariants}
Finally, future work may bypass geometric construction entirely by leveraging tools from Algebraic Geometry and Combinatorics.
  Rather than attacking the full composition, we could do a \textit{Layer-wise Analysis}. The global exponential complexity arises from the composition of layers. A single layer with width $w$ acts as a hyperplane arrangement in $\mathbb{R}^d$, creating $O(w^d)$ regions \cite{Raghu2017}. While still computationally demanding for large $d$, analyzing the topology layer-by-layer decomposes the intractable global problem into a sequence of sub-problems that are more manageable.

We anticipate that future developments in computational algebraic geometry will provide more powerful tools to estimate the $f$-vector and other topological descriptors of these structures without requiring the explicit construction of the full polyhedral complex.

\end{document}